\documentclass{article}
\usepackage{spconf,amsmath,graphicx}
\usepackage{amssymb}
\usepackage{algorithm}
\usepackage{cite}
\usepackage{algorithmic}
\usepackage{multirow}
\usepackage{url}
\usepackage{color}

\title{Group Sparsity Residual with Non-Local Samples for Image Denoising}
\name{ {Zhiyuan~Zha$^{1, \dagger}$, Xinggan~Zhang$^{1}$, Qiong~Wang$^{1,*}$, Yechao~Bai$^{1}$,  Lan~Tang$^{1, 2}$ and Xin~Yuan$^{3, \dagger}$ }
\thanks{$\dagger$ indicates equal contributions.}
\thanks{* Corresponding Author. This work was supported by the NSFC (61571220, 61462052, 61502226).}
 }

\address{  {$^1${School of Electronic Science and Engineering, Nanjing University, Nanjing 210023, China.}} \\
       {$^2${National Mobile Commun. Research Lab., Southeast University, Nanjing 210023, China.}}\\
       {$^3${Nokia Bell Labs, 600 Mountain Avenue, Murray Hill, NJ, 07974, USA.}}}

\begin{document}
\ninept

\maketitle
\begin{abstract}
Inspired by group-based sparse coding, recently proposed group sparsity residual (GSR) scheme has demonstrated superior performance in image processing. However, one challenge in GSR is to estimate the residual by using a proper reference of the group-based sparse coding (GSC), which is desired to be as close to the truth as possible. Previous researches utilized the estimations from other algorithms (i.e., GMM or BM3D), which are either not accurate or too slow. In this paper, we propose to use the Non-Local Samples (NLS) as reference in the GSR regime for image denoising, thus termed GSR-NLS. More specifically, we  first obtain a good estimation of the group sparse coefficients by the image nonlocal self-similarity, and then solve the GSR model by an effective iterative shrinkage algorithm. Experimental results demonstrate that the proposed GSR-NLS not only outperforms many state-of-the-art methods, but also delivers the competitive advantage of speed.
\end{abstract}
\begin{keywords}
Image denoising, group-based sparse coding, group sparsity residual, nonlocal self-similarity, iterative shrinkage algorithm.
\end{keywords}
\section{Introduction}
\label{sec:intro}
Image denoising plays an important role in various image processing tasks.
In general, image denoising aims to restore the clean image $\textbf{\emph{y}}$ from its corrupted observation $\textbf{\emph{z}}=\textbf{\emph{y}}+\textbf{\emph{v}}$, where $\textbf{\emph{v}}$ is usually considered to be an additive white Gaussian noise.
Image denoising problem is mathematically ill-posed and priors are thus usually employed to achieve good results.
Over the past few years, numerous image prior models have been developed, including total variation based \cite{1,2}, wavelet/curvelet based \cite{3,4,40,5}, sparse coding based \cite{6,7}, nonlocal self-similarity based \cite{8,9,12}, and deep learning based~\cite{10,11} ones.


One significant advance in image processing is to model the prior on patches, and a
representative research is sparse coding~\cite{6,7}, which assumes that each patch of an image can be precisely modeled by a sparse linear combination of some fixed and trainable basis elements, which are therefore called atoms and these atoms compose a dictionary. The seminal work of K-SVD dictionary learning method \cite{7} has not only shown promising denoising performance, but also been extended to other image processing and computer vision tasks \cite{13,14}.
Meanwhile,
since image patches with similar structures can be spatially far from each other and thus can be collected across the whole image, the so-called nonlocal self-similarity (NSS) prior is among the most remarkable priors for image restoration \cite{9,15,16,17,18,19,20,21,22,23,24}. The seminal work of nonlocal means (NLM) \cite{8} exploited the NSS prior to carry out a form of the weighted filtering for image denoising. Compared with the local regularization methods (e.g, total variation based method \cite{1}), the nonlocal regularization based methods can retain the image edges and the sharpness effectively.
Inspired by the success of the NSS prior, group-based sparse coding (GSC) has attracted considerable interests in image denoising \cite{17,18,21,22,23,24}. However, due to the influence of noise, the conventional GSC model is accurate enough to restore the original image.

Most recently, the group sparsity residual~\cite{31,32} has been proposed for image denoising, which adopts a reference in each iteration to approximate the true sparse coefficients of each group. The reference was estimated by Gaussian mixture model (GMM) \cite{34,22} or other algorithms, which is either not accurate or too slow.
%
In this paper, we propose a new method for image denoising via group sparsity residual scheme with non-local samples (GSR-NLS). We first obtain a good estimation of the group sparse coefficients of the original image by the image nonlocal self-similarity, and then the group sparse coefficients of the noisy image are inferred to approximate this estimation.
Moreover, we develop an effective iterative shrinkage algorithm to solve the proposed GSR-NLS model. Experimental results show that the proposed GSR-NLS not only outperforms many state-of-the-art methods in terms of the objective and the perceptual metrics, but also deliveries  a competitive speed.

\section{Group-based Sparse Coding for Image Denoising}
\label{sec:2}
In this section, we will briefly introduce the conventional group-based sparse coding (GSC) model for image denoising. Specifically, taking a clean image $\textbf{\emph{y}}\in {\mathbb R}^{\sqrt{N}\times\sqrt{N}}$ as an example, it is divided into $n$ overlapped patches of size $\sqrt{b} \times \sqrt{b}$, and each patch is denoted by a vector $\textbf{\emph{y}}_i\in {\mathbb R}^{b}$, $i=1, 2, ...n$. Then for each patch $\textbf{\emph{y}}_i$, its $m$ similar patches are selected from a searching window with $W \times W$ pixels to form a set ${\cal S}_i$. Following this, all patches in ${\cal S}_i$ are stacked into a matrix $\textbf{\emph{Y}}_i\in {\mathbb R}^{b\times m}$, i.e., $\textbf{\emph{Y}}_i =\{\textbf{\emph{y}}_{i,1}, \textbf{\emph{y}}_{i,2}, ..., \textbf{\emph{y}}_{i,m}\}$. The matrix $\textbf{\emph{Y}}_i$ consisting of patches with similar structures is thereby called a group, where $\{\textbf{\emph{y}}_{i,j}\}_{j=1}^m$ denotes the $j$-th patch in the $i$-th group. Following this, similar to patch-based sparse coding \cite{6,7}, given a dictionary $\textbf{\emph{D}}_i$, each group $\textbf{\emph{Y}}_i$ can be sparsely represented and solved by the following minimization problem,
\begin{equation}
\textstyle{ {{\textbf{\emph{B}}}_i}=\arg\min_{{\textbf{\emph{B}}}_i} \left(||{\textbf{\emph{Y}}}_i-{\textbf{\emph{D}}}_i{{\textbf{\emph{B}}}_i}||_F^2+\lambda||{{\textbf{\emph{B}}}_i}||_1 \right)},
\label{eq:1}
\end{equation} 
where $\lambda$ is the regularization parameter; $||~||_F^2$ denotes the Frobenious norm. Here, $\ell_1$-norm is imposed on each column of ${{\textbf{\emph{B}}}_i}$, which also holds true for the following derivation with  $\ell_1$-norm on matrix.

In image denoising, each patch $\textbf{\emph{z}}_i$ is extracted from the {\em noisy} image $\textbf{\emph{z}}$, and we search for its $m$ similar patches to generate a noisy image patch group $\textbf{\emph{Z}}_i\in {\mathbb R}^{b\times m}$, i.e., $\textbf{\emph{Z}}_i =\{\textbf{\emph{z}}_{i,1}, \textbf{\emph{z}}_{i,2}, ..., \textbf{\emph{z}}_{i,m}\}$. Then, image denoising is translated into how to restore the $\textbf{\emph{Y}}_i$ from $\textbf{\emph{Z}}_i$ using the GSC model,
\begin{equation}
\textstyle{ {{\textbf{\emph{A}}}_i}=\arg\min_{{\textbf{\emph{A}}}_i} \left(||{\textbf{\emph{Z}}}_i-{\textbf{\emph{D}}}_i{{\textbf{\emph{A}}}_i}||_F^2+\lambda||{{\textbf{\emph{A}}}_i}||_1 \right) }.
\label{eq:2}
\end{equation} 
Once all group sparse codes $\{\textbf{\emph{A}}_i\}_{i=1}^n$ are obtained, the underlying clean image $\hat{\textbf{\emph{y}}}$ can be reconstructed as $\hat{\textbf{\emph{y}}}= \textbf{\emph{D}}\textbf{\emph{A}}$, where $\textbf{\emph{A}}$ denotes the set of $\{\textbf{\emph{A}}_i\}_{i=1}^n$.
However, due to the noise, the conventional GSC mode of Eq.~\eqref{eq:2} cannot recover the underlying image ${\textbf{\emph{y}}}$ accurately.

\section{Denoising via Group Sparsity Residual with Non-Local Samples}
\label{sec:3}
Recalling Eq.~\eqref{eq:1} and Eq.~\eqref{eq:2}, the noisy group sparse coefficient $\textbf{\emph{A}}_i$ obtained by solving Eq.~\eqref{eq:2} is excepted to be as close to  the true group sparse coefficient $\textbf{\emph{B}}_i$ of the original image $\textbf{\emph{y}}$ in Eq.~\eqref{eq:1} as possible. Accordingly, the quality of image denoising largely depends on the group sparsity residual, which is defined as the difference between the noisy group sparse coefficient $\textbf{\emph{A}}_i$  and the true group sparse coefficient $\textbf{\emph{B}}_i$,
\begin{equation}
\textstyle{ {{\textbf{\emph{R}}}_i}={{\textbf{\emph{A}}}_i}-{{\textbf{\emph{B}}}_i}} .
\label{eq:3}
\end{equation} 

In order to obtain a good performance in image denoising, we hope that the group sparsity residual ${{\textbf{\emph{R}}}_i}$ of each group is as small as possible. To this end, to reduce the group sparsity residual ${{\textbf{\emph{R}}}}=\{{{\textbf{\emph{R}}}_i}\}_{i=1}^n$ and boost the accuracy of ${{\textbf{\emph{A}}}}$, we propose the group sparsity residual (GSR) based model  for image denoising. This is
\begin{equation}
\textstyle{ {{\textbf{\emph{A}}}_i}=\arg\min_{{\textbf{\emph{A}}}_i} \left(||{\textbf{\emph{Z}}}_i-{\textbf{\emph{D}}}_i{{\textbf{\emph{A}}}_i}||_F^2+\lambda||{{\textbf{\emph{A}}}_i}-{{\textbf{\emph{B}}}_i}||_p \right)},
\label{eq:4}
\end{equation} 

Similarly, $\ell_p$-norm is imposed on each column of ${{\textbf{\emph{R}}}_i}$, which also holds true for the following derivation with  $\ell_p$-norm on matrix. Since the original image $\textbf{\emph{y}}$ is not available, it is impossible to get the true sparse coefficient ${{\textbf{\emph{B}}}_i}$. We will describe how to {\em estimate} ${{\textbf{\emph{B}}}_i}$ and $p$ below.

\begin{figure}[!htbp]
		\centering
		{\includegraphics[width=0.48\textwidth]{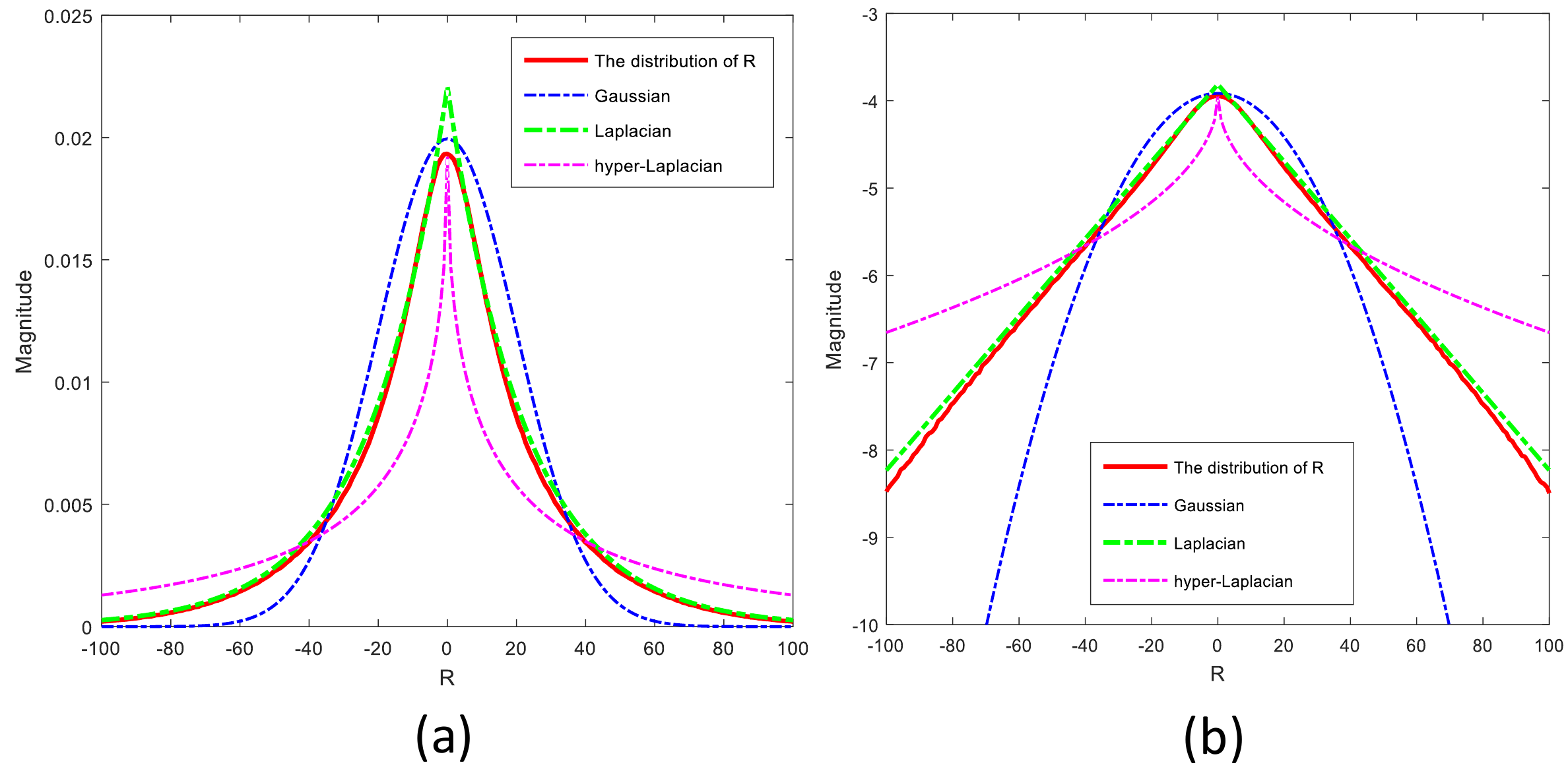}}
\vspace{-5mm}
	\caption{The distribution of the group sparsity residual ${{\textbf{\emph{R}}}}$ for image $\emph{House}$ with $\sigma$=30 and fitting Gaussian, Laplacian and hyper-Laplacian distribution in (a) linear and (b) log domain, respectively.}
	\label{fig:1}
	\vspace{-3mm}
\end{figure}

\subsection {Determine $p$}
\label{subsec:3.1}
Let us come back to Eq.~\eqref{eq:4}, it is clear that one important problem of the proposed GSR based image denoising is the determination of $p$. Here, we conduct some experiments to investigate the statistical property of ${{\textbf{\emph{R}}}}= \{\textbf{\emph{R}}_i\}_{i=1}^n$. We compute ${{\textbf{\emph{A}}}}= \{\textbf{\emph{A}}_i\}_{i=1}^n$ and ${{\textbf{\emph{B}}}}= \{\textbf{\emph{B}}_i\}_{i=1}^n$ by solving Eq.~\eqref{eq:2} and Eq.~\eqref{eq:1}, respectively. The principe component analysis (PCA) based dictionary \cite{15} is used in these experiments. One typical image $\emph{House}$ is used as example, where Gaussian white noise is added with standard deviation $\sigma_n=30$. We plot the empirical distribution of ${{\textbf{\emph{R}}}}$ as well as the fitting Gaussian, Laplacian and hyper-Laplacian distributions in Fig.~\ref{fig:1} (a). To better observe the fitting of the tails, we also plot these distributions in the log domain in Fig.~\ref{fig:1} (b). It can be seen that the empirical distribution of ${{\textbf{\emph{R}}}}$ can be well characterized  by the Laplacian distribution. Therefore, we set $p=1$ and thus the $\ell_1$-norm is adopted to regularize the proposed GSR model,
\begin{equation}
\textstyle{ {{\textbf{\emph{A}}}_i}=\arg\min_{{\textbf{\emph{A}}}_i} \left(||{\textbf{\emph{Z}}}_i-{\textbf{\emph{D}}}_i{{\textbf{\emph{A}}}_i}||_F^2+\lambda||{{\textbf{\emph{A}}}_i}-{{\textbf{\emph{B}}}_i}||_1 \right)}.
\label{eq:5}
\end{equation} 

\subsection {Estimate of the Unknown Group Sparse Coefficient ${{\textbf{\emph{B}}}}$}
\label{subsec:3.2}
\begin{figure*}[!htbp]
	\centerline{\includegraphics[width=18cm]{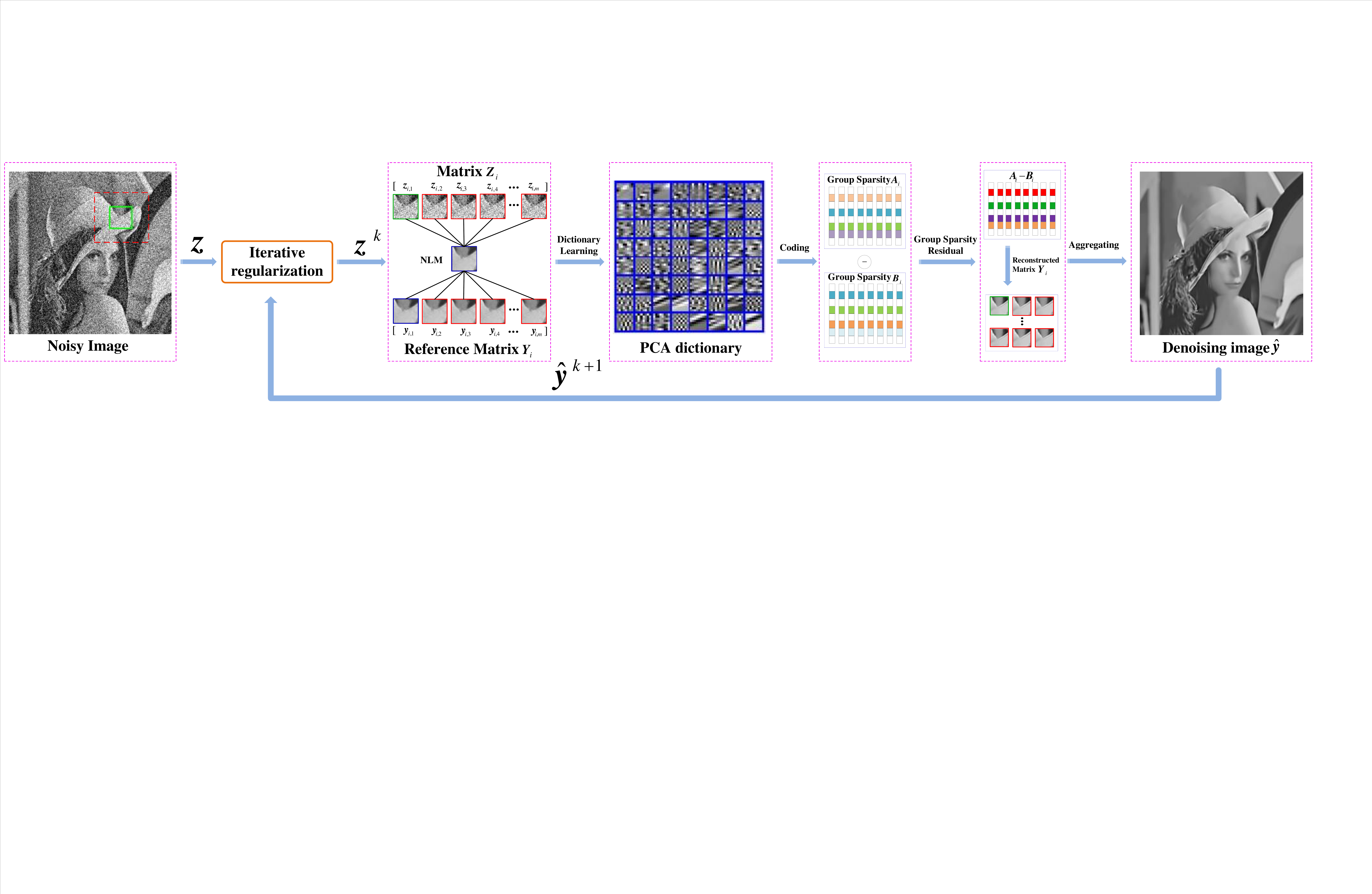}}
	\caption{The flowchart of the proposed GSR-NLS model for image denoising.}
	\label{fig:2}
\end{figure*}

Since the original image $\textbf{\emph{y}}$ is not available in real applications, the true group sparse coefficient $\textbf{\emph{B}}_i$ is unknown. Thus, we need to {\em estimate} $\textbf{\emph{B}}_i$ in Eq.~\eqref{eq:5}. In general, there are  a variety of methods to estimate $\textbf{\emph{B}}_i$, which depends on the prior knowledge of the original image $\textbf{\emph{y}}$. Different from~\cite{31,32}, in this paper, based on the fact that natural images often contain repetitive structures \cite{28}, we search nonlocal similar patches (i.e., non-local samples) to the given patch and use the method similar to nonlocal means \cite{8} to estimate $\textbf{\emph{B}}_i$. Specifically, a good estimation of $\textbf{\emph{b}}_{i, 1}$ can be computed by the weighted average of each element $\textbf{\emph{a}}_{i, j}$ in $\textbf{\emph{A}}_i$ associated with each group including $m$ nonlocal similar patches, where $\textbf{\emph{b}}_{i, 1}$ and $\textbf{\emph{a}}_{i, j}$ represent the first and the $j$-th element of $\textbf{\emph{B}}_i$ and $\textbf{\emph{A}}_i$, respectively. Then we have,
\begin{equation}
\textstyle{ \textbf{\emph{b}}_{i, 1}=\sum\nolimits_{j=1}^m w_{i,j}\textbf{\emph{a}}_{i, j} },
\label{eq:6}
\end{equation} 
where $w_{i,j}$ is the weight, which is inversely proportional to the distance between patches $\textbf{\emph{z}}_{i}$ and $\textbf{\emph{z}}_{i, j}$: $w_{i,j}= {\rm exp}(-||\textbf{\emph{z}}_{i}-\textbf{\emph{z}}_{i, j}||_2^2/h/L)$, where $h$ is a predefined constant and $L$ is a normalization factor \cite{8}.
After this, we simply copy $\textbf{\emph{b}}_{i, 1}$ by $m$ times to estimate $\textbf{\emph{B}}_i$, i.e.,
\begin{equation}
\textstyle{ \textbf{\emph{B}}_i =\{\textbf{\emph{b}}_{i,1}, \textbf{\emph{b}}_{i,2}, ..., \textbf{\emph{b}}_{i,m}\}},
\label{eq:7}
\end{equation} 
where $\{\textbf{\emph{b}}_{i,j}\}_{j=1}^m$ denotes the $j$-th element in the $i$-th group sparse coefficient $\textbf{\emph{B}}_i$ and they are the same. The flowchart of the proposed GSR-NLS model for image denoising is illustrated in Fig.~\ref{fig:2}. Another important issue of the proposed GSR-NLS based image denoising is the selection of the dictionary. To adapt to the local image structures, instead of learning an over-complete dictionary for each group $\textbf{\emph{Z}}_i$ as  in \cite{23}, we utilized the PCA based dictionary \cite{15}.

\subsection {Iterative Shrinkage Algorithm to Solve the Proposed GSR-NLS Model}
\label{subsec:3.3}
Due to orthogonality (obtained by PCA) of each dictionary $\textbf{\emph{D}}_i$, Eq.~\eqref{eq:5} can be rewritten as
\begin{equation}
\begin{aligned}
  {{\textbf{\emph{A}}}_i}& = \textstyle{ \arg\min_{{\textbf{\emph{A}}}_i} \{||{\textbf{\emph{G}}}_i-{{\textbf{\emph{A}}}_i}||_F^2+\lambda||{{\textbf{\emph{A}}}_i}-{{\textbf{\emph{B}}}_i}||_1\}}\\
& =\textstyle{ \arg\min_{{\boldsymbol\alpha}_i} \{||{{\textbf{\emph{g}}}}_i-{{\boldsymbol\alpha}_i}||_2^2+\lambda||{{\boldsymbol\alpha}_i}-{{\boldsymbol\beta}_i}||_1\}},
\end{aligned}
\label{eq:8}
\end{equation} 
where $\textbf{\emph{Z}}_i=\textbf{\emph{D}}_i\textbf{\emph{G}}_i$; ${{\boldsymbol\alpha}_i}$, ${{\boldsymbol\beta}_i}$ and ${{\textbf{\emph{g}}}}_i$ denote the vectorization of the matrix ${{\textbf{\emph{A}}}_i}$, ${{\textbf{\emph{B}}}_i}$ and ${{\textbf{\emph{G}}}_i}$, respectively.

To solve Eq.~\eqref{eq:8} effectively, an iterative shrinkage algorithm \cite{29} is adopted. To be concrete, for fixed ${{\textbf{\emph{g}}}}_i$, ${{\boldsymbol\beta}_i}$ and $\lambda$, we have
 \begin{equation}
\textstyle{ \hat{{\boldsymbol\alpha}}_i = {\rm soft} ({{\textbf{\emph{g}}}}_i-{{\boldsymbol\beta}_i}, \lambda) + {{\boldsymbol\beta}_i}},
\label{eq:9}
\end{equation} 
where {\rm soft} ($\cdot$) is the soft-thresholding operator~\cite{29}.

Following this, the latent clean patch group $\textbf{\emph{Y}}_i$ can be calculated by $\hat{\textbf{\emph{Y}}}_i= \textbf{\emph{D}}_i\hat{\textbf{\emph{A}}}_i$. After obtaining the estimate of all groups $\{\hat{\textbf{\emph{Y}}}_i\}$, we get the full image $\hat{\textbf{\emph{y}}}$ by putting the groups back to their original locations and averaging the overlapped pixels. Moreover, we could execute the above denoising procedures several iterations for better results. In the $k$-th iteration, the iterative regularization strategy \cite{2} is utilized to update the estimation of the noise variance, and thus updating $\textbf{\emph{z}}^k$. The  standard deviation of noise in $k$-th iteration is adjusted as ${\sigma_n}^k=\eta\sqrt{{\sigma_n}^2-||\textbf{\emph{z}}-\textbf{\emph{y}}^{k-1}||_2^2}$, where $\eta$ is a constant.

The parameter $\lambda$ that balances the fidelity term and the regularization term should be adaptively determined in each iteration. Inspired by \cite{3}, the regularization parameter $\lambda$ of each noisy group $\textbf{\emph{Z}}_i$ is set to $\lambda=c* 2\sqrt{2}{\sigma_n}^2/(\delta_i+\varepsilon)$, where $\delta_i$ denotes the estimated variance of $\textbf{\emph{R}}_i$ and $c, \varepsilon$ are small constants.

Throughout the numerical experiments, we choose the following stoping iteration for the proposed denoising algorithm, i.e, $||\hat{\textbf{\emph{y}}}^k-\hat{\textbf{\emph{y}}}^{k-1}||_2^2/||\hat{\textbf{\emph{y}}}^{k-1}||_2^2<\tau$, where $\tau$ is a small constant. The complete description of the proposed GSR-NLS for image denoising is exhibited in Algorithm~\ref{algo:1}.
\begin{center}
	\begin{algorithm}[htbp]
		\caption{The Proposed GSR-NLS for Image Denoising.}
		\begin{algorithmic}[1]
			\REQUIRE Noisy image $\textbf{\emph{z}}$.
			\STATE  Initialize $\hat{\textbf{\emph{y}}}^{0}=\textbf{\emph{z}}$, $\textbf{\emph{z}}^{0}=\textbf{\emph{z}}$, $\boldsymbol\sigma_n$, $b$, $c$, $m$, $h$, $W$, $\gamma$, $\eta$, $\tau$ and $\varepsilon$.
			\FOR{$k=1$ \TO Max-Iter }
			\STATE Iterative Regularization $\textbf{\emph{z}}^{k}=\hat{\textbf{\emph{y}}}^{k-1}+\gamma (\textbf{\emph{z}}-\textbf{\emph{z}}^{k-1})$.
			\FOR{Each patch $\textbf{\emph{z}}_i$ in $\textbf{\emph{z}}^{k}$}	
			\STATE Find nonlocal similar patches to form a group ${\textbf{\emph{Z}}}_i$.
			\STATE Constructing dictionary ${\textbf{\emph{D}}}_i$ by ${\textbf{\emph{Z}}}_i$ using PCA.
			\STATE Update  ${\textbf{\emph{A}}}_i$ computing by ${\textbf{\emph{A}}}_i= {\textbf{\emph{D}}}_i^{-1}{\textbf{\emph{Z}}}_i$.
			\STATE Estimate  ${\textbf{\emph{B}}}_i$ by Eq.~\eqref{eq:6} and Eq.~\eqref{eq:7}.
			\STATE Update $\lambda$ computing by $\lambda=c*2\sqrt{2}{\sigma_n}^2/(\delta_i+\varepsilon)$.
			\STATE Estimate  $\hat{{\textbf{\emph{A}}}}_i$ by  Eq.~\eqref{eq:9}.
			\STATE Get the estimation:   $\hat{\textbf{\emph{Y}}}_i=\textbf{\emph{D}}_i\hat{{\textbf{\emph{A}}}}_i$.
			\ENDFOR
			\STATE Aggregate ${\textbf{\emph{Y}}}_i$  to form the denoised image $\hat{\textbf{\emph{y}}}^{k}$.
			\ENDFOR
			\STATE $\textbf{Output:}$ The final denoised image $\hat{\textbf{\emph{y}}}$.
		\end{algorithmic}
		\label{algo:1}
	\end{algorithm}
\end{center}

\subsection{Different from Existing Methods}
\label{sec:4}
Now we discuss the difference between the proposed GSR-NLS method, and NCSR \cite{30} and our previous approaches \cite{31,32}.

The main difference between NCSR \cite{30} and the proposed GSR-NLS is that NCSR is essentially a  patch-based sparse coding method, which usually ignored the relationship among similar patches \cite{17,21,23,33}. In addition, NCSR extracted image patches from noisy image $\textbf{\emph{z}}^{k}$ and used $K$-means algorithm to generate $K$ clusters. Following this, it learned $K$ PCA sub-dictionaries from each cluster. However, since each cluster includes thousands of patches, the dictionary learned by PCA from each cluster may not accurately centralize the image features. The proposed GSR-NLS learned the PCA dictionary from each group and the patches in each group are similar, and therefore, our PCA dictionary is more appropriate. An additional advantage of GSR-NLS is  that it only requires  1/3 computational time of NCSR but achieves $\sim$0.5dB improvement on average over NCSR (see Section~\ref{sec:5} for details).

Our previous work in \cite{31} utilized the pre-filtering BM3D \cite{9} to estimate $\textbf{\emph{B}}$, and thus the denoising performance largely depends on the pre-filtering. In other words, if the pre-filtering cannot be used, the denoising method will fail.
Our another work \cite{32} estimated $\textbf{\emph{B}}$ from example image set based on  GMM  \cite{34}. However, under many practical situations, the example image set is simply unavailable. Moreover, the computation speed of this method is very slow.
Therefore, the proposed GSR-NLS is more reasonable and generic while providing very similar results as in~\cite{31,32}.

\begin{figure}[!htbp]
	\begin{minipage}[b]{1\linewidth}
		\centering
		\centerline{\includegraphics[width=8.5cm]{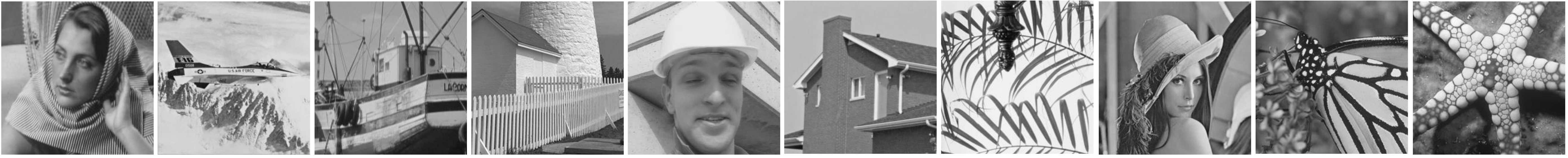}}
	\end{minipage}
	\vspace{-5mm}
	\caption{The test images for denoising experiments. }
	\label{fig:3}
\end{figure}

	\begin{table*}[!htbp]
		\caption{PSNR ($\textnormal{d}$B)  results of different denoising methods.}
		\centering
		\resizebox{1.00\textwidth}{!}
		{
			\begin{tabular}{|c|c|c|c|c|c|c|c|c|c|c||c|c|c|c|c|c|c|c|c|c|c|c|c|c|}
				\hline
				\multicolumn{1}{|c|}{}&\multicolumn{10}{|c||}{$\sigma_n=20$}&\multicolumn{10}{|c|}{$\sigma_n=40$}\\
				\hline
				\multirow{2}{*}{\textbf{{Images}}}&\multirow{2}{*}{\textbf{{BM3D}}}
				&\multirow{2}{*}{\textbf{{EPLL}}}&\multirow{2}{*}{\textbf{{Plow}}}&\multirow{2}{*}{\textbf{{NCSR}}}&\multirow{2}{*}{\textbf{{PID}}}
				&\multirow{2}{*}{\textbf{{PGPD}}}&\multirow{2}{*}{\textbf{{aGMM}}}&\multirow{2}{*}{\textbf{{LINC}}}&{\textbf{{AST-}}}&{\textbf{{GSR-}}}
                &\multirow{2}{*}{\textbf{{BM3D}}}&\multirow{2}{*}{\textbf{{EPLL}}}&\multirow{2}{*}{\textbf{{NCSR}}}&\multirow{2}{*}{\textbf{{PID}}}&\multirow{2}{*}{\textbf{{PGPD}}} &\multirow{2}{*}{\textbf{{PGPD}}}&\multirow{2}{*}{\textbf{{aGMM}}}&\multirow{2}{*}{\textbf{{LINC}}}&{\textbf{{AST-}}}&{\textbf{{GSR-}}}\\
				& &  & & & & & & &{\textbf{NLS}} &{\textbf{NLS}} & & & & & & & & & {\textbf{NLS}}& {\textbf{NLS}}  \\
				\hline
				\multirow{1}{*}{Airplane}
		        &	30.59 	&	30.60 	&	29.98 	&	30.50 	&	30.71 	&	30.80 	&	30.54 	&	30.57 	&	30.70 	&	\textbf{30.87}
                &	26.88 	&	27.08 	&	26.70 	&	26.78 	&	\textbf{27.25} 	&	27.12 	&	26.95 	&	27.08 	&	27.10 	&	27.21
     			\\
				\hline
				\multirow{1}{*}{Barbara}
		        &	31.24 	&	29.85 	&	30.75 	&	31.10 	&	30.98 	&	31.12 	&	30.51 	&	\textbf{31.70} 	&	31.43 	&	31.47
                &	27.26 	&	25.99 	&	27.59 	&	27.25 	&	27.68 	&	27.43 	&	26.34 	&	27.77 	&	27.41 	&	\textbf{27.85}
     			\\
				\hline
				\multirow{1}{*}{boats}
		        &	31.42 	&	30.87 	&	30.90 	&	31.26 	&	31.27 	&	31.38 	&	31.20 	&	31.52 	&	31.50 	&	\textbf{31.55}
                &	27.76 	&	27.42 	&	27.55 	&	27.52 	&	27.73 	&	27.90 	&	27.60 	&	27.86 	&	27.80 	&	\textbf{27.97}
     			\\
				\hline
				\multirow{1}{*}{Fence}
		        &	29.93 	&	29.24 	&	29.13 	&	30.05 	&	30.01 	&	29.99 	&	29.46 	&	30.08 	&	\textbf{30.28} 	&	30.20
                &	26.84 	&	25.74 	&	26.42 	&	26.76 	&	26.94 	&	26.91 	&	25.80 	&	27.07 	&	27.11 	&	\textbf{27.16}
     			\\
				\hline
				\multirow{1}{*}{foreman}
		        &	34.54 	&	33.67 	&	34.21 	&	34.42 	&	34.65 	&	34.44 	&	34.20 	&	\textbf{34.76} 	&	34.55 	&	34.67
                &	31.29 	&	30.28 	&	30.90 	&	31.52 	&	31.81 	&	31.55 	&	30.95 	&	31.31 	&	31.29 	&	\textbf{31.81}
     			\\
				\hline
				\multirow{1}{*}{House}
		        &	33.77 	&	32.99 	&	33.40 	&	33.81 	&	33.70 	&	33.85 	&	33.52 	&	33.82 	&	33.87 	&	\textbf{33.91}
                &	30.65 	&	29.89 	&	30.25 	&	30.79 	&	30.76 	&	31.02 	&	30.40 	&	31.00 	&	30.91 	&	\textbf{31.16}
     			\\
				\hline
				\multirow{1}{*}{Leaves}
		        &	30.09 	&	29.40 	&	29.08 	&	30.34 	&	30.13 	&	30.46 	&	30.05 	&	30.24 	&	30.72 	&	\textbf{30.94}
                &	25.69 	&	25.62 	&	25.45 	&	26.20 	&	26.26 	&	26.29 	&	25.76 	&	26.31 	&	26.69 	&	\textbf{26.82}
     			\\
				\hline
				\multirow{1}{*}{Lena}
		        &	31.52 	&	31.25 	&	30.98 	&	31.48 	&	31.57 	&	31.64 	&	31.48 	&	\textbf{31.80} 	&	31.63 	&	31.71
                &	27.82 	&	27.78 	&	27.78 	&	28.00 	&	28.18 	&	\textbf{28.22} 	&	27.91 	&	28.13 	&	28.00 	&	28.16
     			\\
				\hline	
				\multirow{1}{*}{Monarch}
		        &	30.35 	&	30.49 	&	29.50 	&	30.52 	&	30.59 	&	30.68 	&	30.31 	&	30.64 	&	30.84 	&	\textbf{30.98}
                &	26.72 	&	26.89 	&	26.43 	&	26.81 	&	27.27 	&	27.02 	&	26.87 	&	27.14 	&	27.20 	&	\textbf{27.33}
     			\\
				\hline		
				\multirow{1}{*}{starfish}
		        &	29.67 	&	29.58 	&	28.83 	&	29.85 	&	29.36 	&	29.84 	&	29.74 	&	29.58 	&	30.04 	&	\textbf{30.08}
                &	26.06 	&	26.12 	&	25.70 	&	26.17 	&	25.92 	&	26.21 	&	26.16 	&	26.07 	&	26.36 	&	\textbf{26.53}
     			\\
				\hline		
				\multirow{1}{*}{\textbf{Average}}
		        &	31.31 	&	30.79 	&	30.67 	&	31.34 	&	31.30 	&	31.42 	&	31.10 	&	31.47 	&	31.56 	&	\textbf{31.64}
                &	27.70 	&	27.28 	&	27.48 	&	27.78 	&	27.98 	&	27.97 	&	27.48 	&	27.97 	&	27.99 	&	\textbf{28.20}
     			\\
				\hline	
                \multicolumn{1}{|c|}{}&\multicolumn{10}{|c||}{$\sigma_n=50$}&\multicolumn{10}{|c|}{$\sigma_n=75$}\\
				\hline
				\multirow{2}{*}{\textbf{{Images}}}&\multirow{2}{*}{\textbf{{BM3D}}}
				&\multirow{2}{*}{\textbf{{EPLL}}}&\multirow{2}{*}{\textbf{{Plow}}}&\multirow{2}{*}{\textbf{{NCSR}}}&\multirow{2}{*}{\textbf{{PID}}}
				&\multirow{2}{*}{\textbf{{PGPD}}}&\multirow{2}{*}{\textbf{{aGMM}}}&\multirow{2}{*}{\textbf{{LINC}}}&{\textbf{{AST-}}}&{\textbf{{GSR-}}}
                &\multirow{2}{*}{\textbf{{BM3D}}}&\multirow{2}{*}{\textbf{{EPLL}}}&\multirow{2}{*}{\textbf{{NCSR}}}&\multirow{2}{*}{\textbf{{PID}}}&\multirow{2}{*}{\textbf{{PGPD}}} &\multirow{2}{*}{\textbf{{PGPD}}}&\multirow{2}{*}{\textbf{{aGMM}}}&\multirow{2}{*}{\textbf{{LINC}}}&{\textbf{{AST-}}}&{\textbf{{GSR-}}}\\
				& &  & & & & & & &{\textbf{NLS}} &{\textbf{NLS}} & & & & & & & & & {\textbf{NLS}}& {\textbf{NLS}}  \\
				\hline
				\multirow{1}{*}{Airplane}
		        &	25.76 	&	25.96 	&	25.64 	&	25.63 	&	26.09 	&	25.98 	&	25.83 	&	26.04 	&	26.02 	&	\textbf{26.17}
                &	23.99 	&	24.03 	&	23.67 	&	23.76 	&	24.08 	&	\textbf{24.15} 	&	23.95 	&	23.81 	&	24.06 	&	24.12
     			\\
 				\hline
 				\multirow{1}{*}{Barbara}
		        &	26.42 	&	24.86 	&	26.42 	&	26.13 	&	\textbf{26.58} 	&	26.27 	&	25.37 	&	26.27 	&	26.43 	&	26.51
                &	24.53 	&	23.00 	&	24.30 	&	24.06 	&	\textbf{24.67} 	&	24.39 	&	23.09 	&	24.03 	&	24.40 	&	24.46
     			\\
 				\hline
  				\multirow{1}{*}{boats}
		        &	26.74 	&	26.31 	&	26.38 	&	26.37 	&	26.58 	&	26.82 	&	26.50 	&	26.70 	&	26.78 	&	\textbf{26.95}
                &	24.82 	&	24.33 	&	24.23 	&	24.44 	&	24.51 	&	24.83 	&	24.51 	&	24.44 	&	24.76 	&	\textbf{24.94}
     			\\
 				\hline
   				\multirow{1}{*}{Fence}
		        &	25.92 	&	24.57 	&	25.49 	&	25.77 	&	25.94 	&	25.94 	&	24.57 	&	25.89 	&	26.22 	&	\textbf{26.26}
                &	24.22 	&	22.46 	&	23.57 	&	23.75 	&	24.20 	&	24.18 	&	22.70 	&	23.81 	&	24.40 	&	\textbf{24.53}
     			\\
 				\hline
    			\multirow{1}{*}{foreman}
		        &	30.36 	&	29.20 	&	29.60 	&	30.41 	&	30.63 	&	30.45 	&	29.80 	&	30.33 	&	30.46 	&	\textbf{30.77}
                &	28.07 	&	27.24 	&	27.15 	&	28.18 	&	28.40 	&	28.39 	&	27.67 	&	28.11 	&	28.54 	&	\textbf{28.75}
     			\\
 				\hline
     			\multirow{1}{*}{House}
		        &	29.69 	&	28.79 	&	28.99 	&	29.61 	&	29.58 	&	29.93 	&	29.28 	&	29.87 	&	30.13 	&	\textbf{30.45}
                &	27.51 	&	26.70 	&	26.52 	&	27.16 	&	27.35 	&	27.81 	&	27.11 	&	27.56 	&	28.06 	&	\textbf{28.59}
     			\\
 				\hline
      			\multirow{1}{*}{Leaves}
		        &	24.68 	&	24.39 	&	24.28 	&	24.94 	&	25.01 	&	25.03 	&	24.42 	&	25.11 	&	25.32 	&	\textbf{25.66}
                &	22.49 	&	22.03 	&	22.02 	&	22.60 	&	22.61 	&	22.61 	&	21.96 	&	22.45 	&	22.95 	&	\textbf{23.34}
     			\\
 				\hline
       			\multirow{1}{*}{Lena}
		        &	26.90 	&	26.68 	&	26.70 	&	26.94 	&	27.09 	&	\textbf{27.15} 	&	26.85 	&	26.94 	&	27.08 	&	27.06
                &	25.17 	&	24.75 	&	24.64 	&	25.02 	&	25.16 	&	25.30 	&	25.02 	&	25.12 	&	25.32 	&	\textbf{25.32}
     			\\
 				\hline
        		\multirow{1}{*}{Monarch}
		        &	25.82 	&	25.78 	&	25.41 	&	25.73 	&	26.21 	&	26.00 	&	25.82 	&	25.88 	&	26.12 	&	\textbf{26.25}
                &	23.91 	&	23.73 	&	23.34 	&	23.67 	&	24.22 	&	24.00 	&	23.85 	&	23.91 	&	24.11 	&	\textbf{24.35}
     			\\
 				\hline
         		\multirow{1}{*}{starfish}
		        &	25.04 	&	25.05 	&	24.71 	&	25.06 	&	24.80 	&	25.11 	&	25.09 	&	24.81 	&	25.26 	&	\textbf{25.36}
                &	23.27 	&	23.17 	&	22.82 	&	23.18 	&	22.89 	&	23.23 	&	23.22 	&	22.74 	&	23.24 	&	\textbf{23.32}
     			\\
 				\hline
          		\multirow{1}{*}{\textbf{Average}}
		        &	26.73 	&	26.16 	&	26.36 	&	26.66 	&	26.85 	&	26.87 	&	26.35 	&	26.78 	&	26.98 	&	\textbf{27.14}
                &	24.80 	&	24.15 	&	24.23 	&	24.58 	&	24.81 	&	24.89 	&	24.31 	&	24.60 	&	24.98 	&	\textbf{25.17}
     			\\
 				\hline
			\end{tabular}}
			\label{Tab:1}
		\end{table*}

\section{Experimental Results}
\label{sec:5}
In this section, we report the performance of the proposed GSR-NLS method for image denoising and compare it with several state-of-the-art denoising methods, including BM3D \cite{9}, EPLL \cite{34}, Plow \cite{35}, NCSR \cite{30}, PID \cite{36}, PGPD \cite{22}, aGMM \cite{37}, LINC \cite{38} and AST-NLS \cite{39}. The parameter setting of the proposed GSR-NLS is as follows. The searching window $W \times W$ is set to be 25$\times$25 and $\varepsilon$ is set to 0.2. The size of patch $\sqrt{b} \times \sqrt{b}$ is set to be 6$\times$6, 7$\times$7, 8$\times$8 and 9$\times$9 for $\sigma_n\leq20$, $20<\sigma_n\leq50$, $50<\sigma_n\leq75$ and $75<\sigma_n\leq100$, respectively. The parameters ($c, \eta, \gamma, m, h, \tau$) are set to (0.8, 0.2, 0.5, 60, 45, 0.0003), (0.7, 0.2, 0.6, 60, 45, 0.0008), (0.6, 0.1, 0.6, 60, 60, 0.002), (0.7, 0.1, 0.5, 70, 80, 0.002), (0.7, 0.1, 0.5, 80, 115, 0.001), (0.7, 0.1, 0.5, 90, 160, 0.0005) and (1, 0.1, 0.5, 100, 160, 0.0005) for $\sigma_n\leq10$, $10<\sigma_n\leq20$, $20<\sigma_n\leq30$, $30<\sigma_n\leq40$, $40<\sigma_n\leq50$, $50<\sigma_n\leq75$ and $75<\sigma_n\leq100$, respectively. The test images are displayed in Fig.~\ref{fig:3}. The source code of the proposed GSR-NLS for image denoising can be downloaded at: \textcolor{red}{\textbf{\url{https://drive.google.com/open?id=0B0wKhHwcknCjZkh4NkprbVhBMFk}.}}

\begin{figure}[!htb]
	\begin{minipage}[b]{1\linewidth}
		\centering
		\centerline{\includegraphics[width=\textwidth]{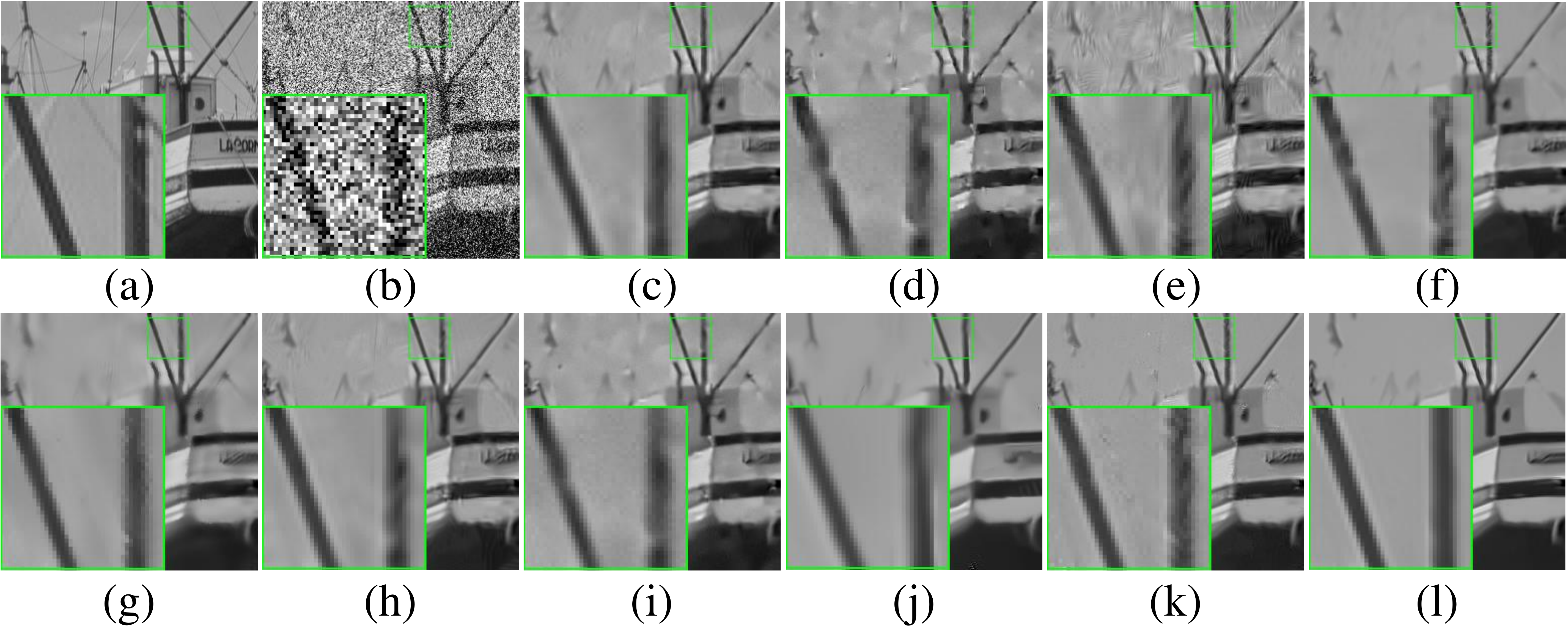}}
	\end{minipage}
	\caption{ {Denoising results on image $\emph{boats}$ by different methods (noise level $\sigma_n$=75). (a) Original image; (b) Noisy image;  (c) BM3D \cite{9} (PSNR=24.82dB); (d) EPLL \cite{34} (PSNR=24.33dB); (e) Plow \cite{35} (PSNR=24.23dB); (f) NCSR \cite{30} (PSNR=24.44dB); (g) PID \cite{36} (PSNR=24.51dB); (h) PGPD \cite{22} (PSNR=24.83dB); (i) aGMM \cite{37} (PSNR= 24.51dB); (j) LINC \cite{38} (PSNR= 24.44dB); (k) AST-NLS \cite{39} (PSNR= 24.76dB); (l) GSR-NLS (PSNR=\textbf{24.94dB}).}}
	\label{fig:4}
\end{figure}

\begin{figure}[!htb]
	\begin{minipage}[b]{1\linewidth}
		\centering
		\centerline{\includegraphics[width=\textwidth]{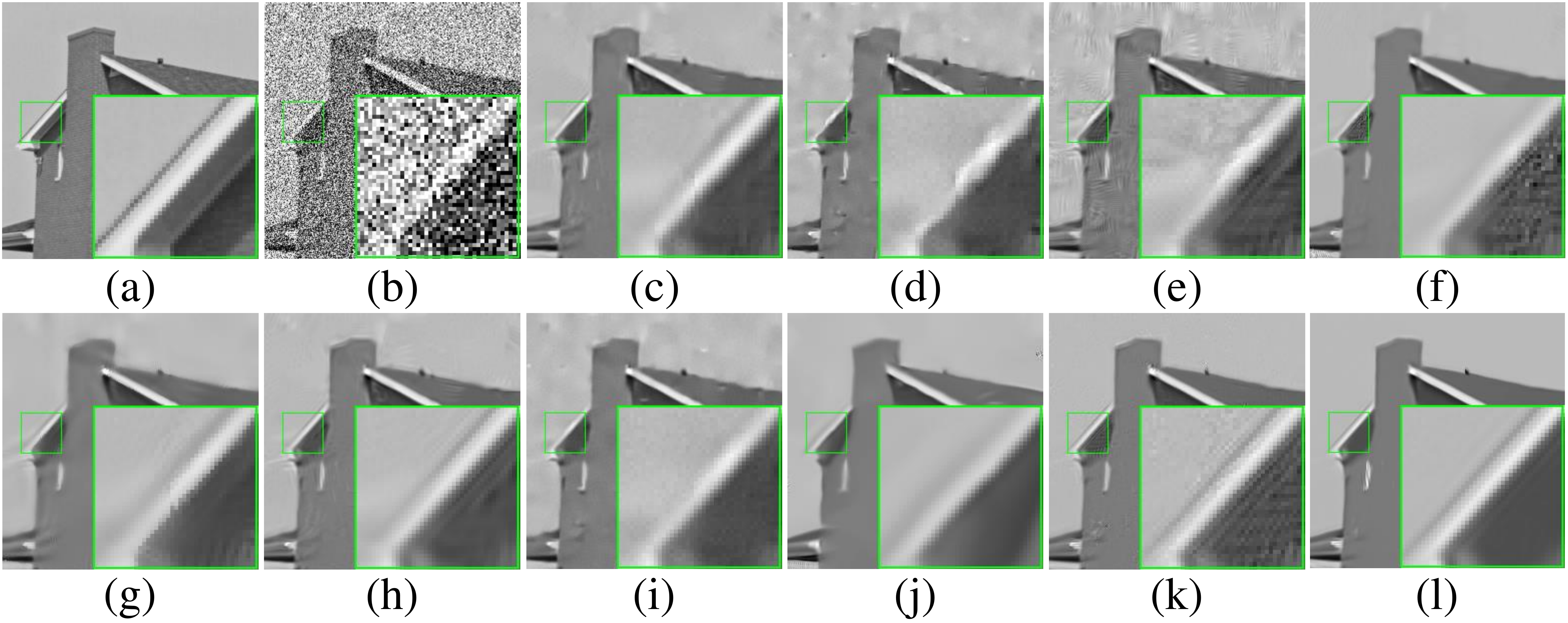}}
	\end{minipage}
	\caption{ {Denoising results on image $\emph{House}$ by different methods (noise level $\sigma_n$=75). (a) Original image; (b) Noisy image;  (c) BM3D \cite{9} (PSNR=27.51dB); (d) EPLL \cite{34} (PSNR=26.70dB); (e) Plow \cite{35} (PSNR=26.52dB); (f) NCSR \cite{30} (PSNR=27.16dB); (g) PID \cite{36} (PSNR=27.35dB); (h) PGPD \cite{22} (PSNR=27.81dB); (i) aGMM \cite{37} (PSNR= 27.11dB); (j) LINC \cite{38} (PSNR= 27.56dB); (k) AST-NLS \cite{39} (PSNR= 28.06dB); (l) GSR-NLS (PSNR=\textbf{28.59dB}).}}
	\label{fig:5}
\end{figure}

Due to the page limit, we only present the denoising results at four noise levels, i.e., Gaussian white noise with standard deviations $\{\sigma_n=20, 40, 50, 75\}$. As shown in Table~\ref{Tab:1}, the proposed GSR-NLS outperforms the other competing methods in most cases in terms of PSNR. The average gains of the proposed GSR-NLS over BM3D, EPLL, Plow, NCSR, PID, PGPD, aGMM, LINC and AST-NLS are as much as 0.40dB, 0.94dB, 0.85dB, 0.45dB, 0.31dB, 0.25dB, 0.73dB, 0.33dB and 0.16dB, respectively. The visual comparisons of different denoising methods with two example images are shown in Fig.~\ref{fig:4} and Fig.~\ref{fig:5}.  It can be seen that BM3D, PID, LINC and AST-NLS are leading to over-smooth phenomena, while EPLL, Plow, NCSR, PGPD and aGMM are likely to produce some undesirable ringing artifacts. By contrast, the proposed GSR-NLS is able to preserve the image local structures and suppress undesirable ringing artifacts more effectively than the other competing methods.

	\begin{table}[!htbp]
		\vspace{-3mm}
		\caption{\small Average running time (s) on the 10 test images.}
		\centering  
		\resizebox{0.48\textwidth}{!}
		{
			\begin{tabular}{|c|c|c|c|c|c|c|c|c|c|c|}
				\hline
				\multirow{2}{*}{\textbf{{Methods}}}&\multirow{2}{*}{\textbf{{BM3D}}}
				&\multirow{2}{*}{\textbf{{EPLL}}}&\multirow{2}{*}{\textbf{{Plow}}}&\multirow{2}{*}{\textbf{{NCSR}}}&\multirow{2}{*}{\textbf{{PID}}}
				&\multirow{2}{*}{\textbf{{PGPD}}}&\multirow{2}{*}{\textbf{{aGMM}}}&\multirow{2}{*}{\textbf{{LINC}}}&{\textbf{{AST-}}}&{\textbf{{GSR-}}}\\
				& &  & & & & & & &{\textbf{NLS}} &{\textbf{NLS}}  \\
				\hline
				\multirow{1}{*}{\textbf{Time}}
				&	2.99 	&	58.73 	&	274.46 	&	385.13 	&	200.88 	&	11.75 	&	250.58 	&	253.14 	&	391.84 	&	{119.73}
				\\
				\hline
			\end{tabular}}
			\vspace{-5mm}
			\label{Tab:2}
		\end{table}
		
Efficiency is another key factor in evaluating a denoising algorithm. To evaluate the computational cost of the competing algorithms, we compare the running time on 10 test images with different noise levels. All experiments are conducted under the Matlab 2015b environment on a computer with Intel (R) Core (TM) i3-4150 with 3.56Hz CPU and 4GB memory. The average running time (in seconds) of different methods is shown in Table~\ref{Tab:2}. It can be seen that the proposed GSR-NLS uses less computation time than  the competing methods except for BM3D, EPLL and PGPD. However, BM3D is implemented with compiled C++ mex-function and is performed in parallel. EPLL and PGPD are based on learning methods, which require a significant amount of time in learning stage. Therefore, the proposed GSR-NLS enjoys a competitive speed.
\section{Conclusion}
\label{sec:6}
This paper has proposed a new method for image denoising using group sparsity residual scheme with nonlocal samples.
We obtained a good estimation of the group sparse coefficients of the original image from the image nonlocal self-similarity and used it in the group sparsity residual model. An effective iterative shrinkage algorithm has been developed to solve the proposed GSR-NLS model.
Experimental results have demonstrated that the proposed GSR-NLS not only outperforms many state-of-the-art methods, but also leads to a competitive speed.


\end{document}